\documentclass{svmult}

\usepackage{mathptmx}
\usepackage{helvet}
\usepackage{courier}
\usepackage{amsmath,amsfonts}
\usepackage{graphicx}
\usepackage[bottom]{footmisc}

\begin{document}

\title*{Compensation of compliance errors in parallel manipulators composed of non-perfect kinematic chains}
\author{Alexandr Klimchik$^{a,b}$, Anatol Pashkevich$^{a,b}$, Damien Chablat$^{a}$ and Geir Hovland$^{c}$}

\institute{ $^{a}$Institut de Recherche en Communications et Cybernetique de Nantes, France;\\
  $^{b}$Ecole des Mines de Nantes, France;
  $^{c}$University of Agder, Norway;\\
  \email{alexandr.klimchik@mines-nantes.fr, anatol.pashkevich@mines-nantes.fr, \\damienl.chablat@irccyn.ec-nantes.fr, geir.hovland@uia.no}
}

\titlerunning{Compensation of compliance errors in parallel manipulators}
\authorrunning{A. Klimchik, A. Pashkevich, D. Chablat and G. Hovland}
\maketitle

\abstract{%
  The paper is devoted to the compliance errors compensation for parallel manipulators under external loading. Proposed approach is based on the non-linear stiffness modeling and reduces to a proper adjusting of a target trajectory. In contrast to previous works, in addition to compliance errors caused by machining forces, the problem of assembling errors caused by inaccuracy in the kinematic chains is considered. The advantages and practical significance of the proposed approach are illustrated by examples that deal with groove milling with Orthoglide manipulator.}




\section{Introduction}
In many robotic applications such as machining, grinding, trimming etc., the interaction between the workpiece and technological tool causes essential deflections that significantly decrease the processing accuracy and quality of the final product. To overcome this difficulty, it is possible to modify either control algorithm or the prescribed trajectory, which is used as the reference input for a control system [1]. This paper focuses on the second approach that is considered to be more realistic in the practice. In contrast to the previous works, the proposed compliance error compensation technique is based on the non-linear stiffness model of the manipulator that is able to take into account significant external loading [2].
 
Usually, the problem of the robot error compensation can be solved in two ways that differ in degree of modification of the robot control software: 
\begin{enumerate} 
	\item[(a)] by modification of the manipulator model (Fig. 1a) which better suits to the real manipulator and is used by the robot controller (in simple case, it can be limited by tuning of the nominal manipulator model, but may also involve essential model enhancement by introducing additional parameters, if it is allowed by the robot manufacturer); 
	\item[(b)] by modification of the robot control program (Fig. 1b) that defines the prescribed trajectory in Cartesian space (here, using relevant error model, the input trajectory is generated in a such way that under the loading the output trajectory coincides with the desired one, while input trajectory differs from the target one).
\end{enumerate}

\begin{figure}[t]
\center
\includegraphics{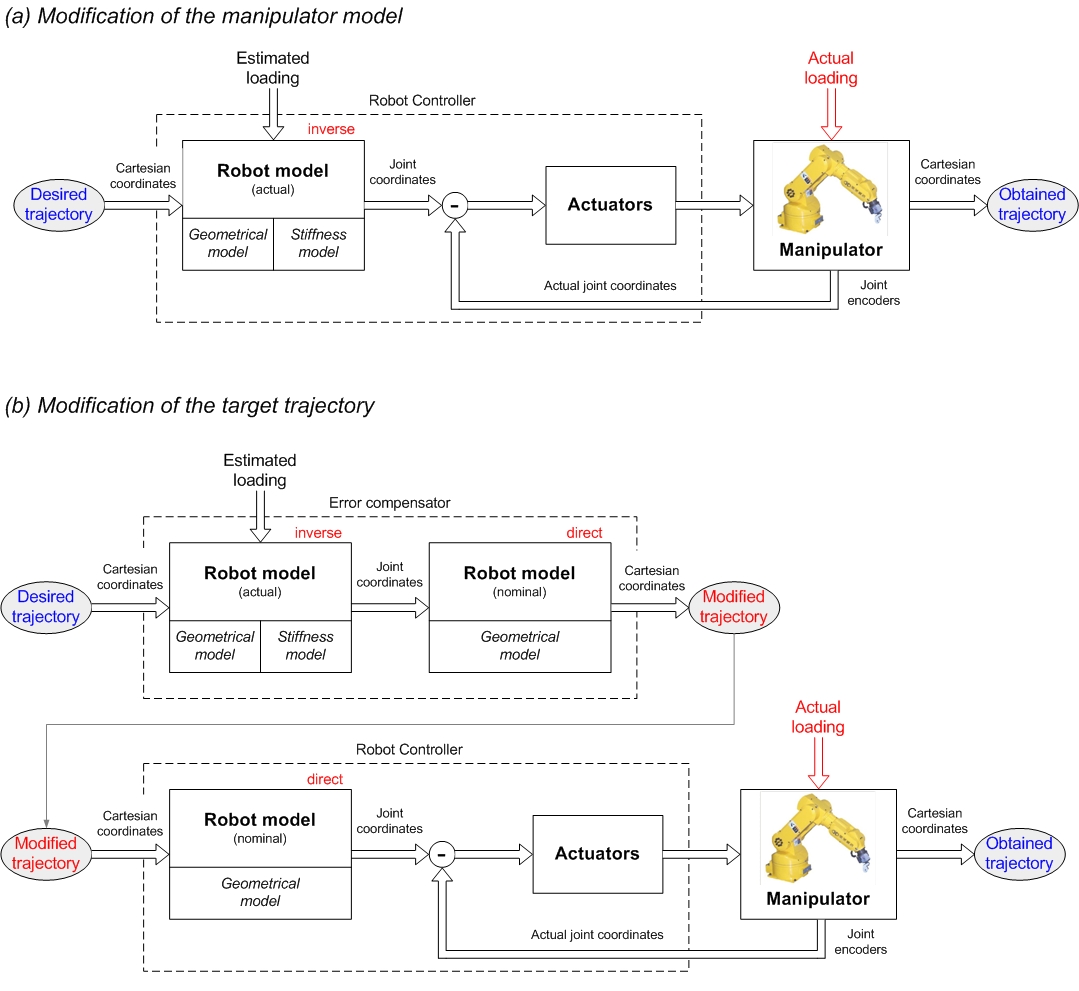}
\caption{Robot error compensation methods.}
\label{Figure:1}
\end{figure} 

It is clear that the first approach can be implemented in on-line mode, while the second one requires preliminary off-line computations. But in practice it is rather unrealistic to include the stiffness model in a commercial robot controller where all transformations between the joint and Cartesian coordinates are based on the manipulator geometrical model. In contrast, the off-line error compensation, based on the second approach, is attractive for industrial applications.

For the geometrical errors, relevant compensation techniques are already well developed. Comprehensive review of related works is given in [3]. In the frame of this work, it is assumed that the geometrical errors are less essential compared to the non-geometrical ones caused by the interaction between the machining tool and workpiece. So, the main attention will be paid to the compliance errors and their compensation techniques.

\section{Problem of compliance error compensation}
For the compliance errors, the compensation technique must rely on two components. The first of them describes distribution of the stiffness properties throughout the workspace and is defined by the stiffness matrix as a function of the joint coordinates or the end-effector location [2]. The second component describes the forces/torques acting on the end-effector while the manipulator is performing its manufacturing task (manipulator loading). In this work, it is assumed that the second component is given and can be obtained either from the dedicated technological process model (that take into account the tool wear, type of machining process, cutting speed, rake angle, cutting fluid, workpiece shape etc) or by direct measurements using the force/torque sensor integrated into the end-effector.

The stiffness matrix required for the compliance errors compensation highly depends on the robot configuration and essentially varies throughout the workspace. From general point of view, full-scale compensation of the compliance errors requires essential revision of the manipulator model embedded in the robot controller. In fact, instead of conventional geometrical model that provides inverse/direct coordinate transformations from the joint to Cartesian spaces and vice versa, here it is necessary to employ the so-called kinetostatic model [4]. It is essentially more complicated than the geometrical model and requires intensive computations. 

If the compliance errors are relatively small, composition of conventional geometrical model and the stiffness matrix give rather accurate approximation of the modified mapping from the joint to Cartesian space. In this case, for the first compensation scheme (see Fig. 1a), the kinetostatic model can be easily implemented on-line if there is an access to the control software modification. Otherwise, the second scheme (see Fig. 1b) can be easily applied. Moreover, with regard to the robot-based machining, there is a solution that does not require force/torque measurements or computations [1] where the target trajectory for the robot controller is modified by applying the "mirror" technique. However, this approach is only suitable for the large-scale production where the manufacturing task and the workpiece location remains the same. Hence, to be applied to the robotic-based machining, the existing compliance errors compensation techniques should be essentially revised to take into account essential forces and torques as well as some other important error sources (inaccuracy in serial chains, for instance). 

\section{Nonlinear technique for compliance error compensation} 
In industrial robotic controllers, the manipulator motions are usually generated using the inverse kinematic model that allows us to compute the input signals for actuators $\mathbf{\rho}_{0}$ corresponding to the desired end-effector location $\mathbf{t}_{0}$, which is assigned assuming that the compliance errors are negligible. However, if the external loading $\mathbf{F'}$ is essential, the kinematic control becomes non-applicable because of changes in the end-effector location. It can be computed from the nonlinear compliance model as

\begin{equation}\label{Eq:1}
{{\mathbf{t}}_{\text{F}}}={{f}^{-1}}\left( \mathbf{F}|{{\mathbf{t}}_{0}} \right)	
\end{equation}	
where the subscripts 'F' and '0' refer to the loaded and unloaded modes respectively, and '$|$ ' separates arguments and parameters of the function $f\left( {} \right)$. Some details concerning this function are given in our previous publication [2]. It should be mentioned that function (1) takes into account loop-closure constraints and validates both for serial and parallel manipulators.

To compensate this undeterred end-effector displacement from ${{\mathbf{t}}_{0}}$ to ${{\mathbf{t}}_{\text{F}}}$, the target point should be modified in a such way that, under the loading $\mathbf{F}$, the end-effector is located in the desired point ${{\mathbf{t}}_{0}}$. This requirement can be expressed using the stiffness model in the following way

\begin{equation}\label{Eq:2}
\mathbf{F}=f\left( {{\mathbf{t}}_{0}}|\mathbf{t}_{\text{0}}^{\text{(F)}} \right)
\end{equation}
where $\mathbf{t}_{\text{0}}^{\text{(F)}}$ denotes the modified target location. Hence, the problem is reduced to the solution of the nonlinear equation (2) for $\mathbf{t}_{\text{0}}^{\text{(F)}}$, while $\mathbf{F}$ and ${{\mathbf{t}}_{0}}$ are assumed to be given. It is worth mentioning that this equation completely differs from the equation $\mathbf{F}=f(\mathbf{t}|{{\mathbf{t}}_{0}})$, where the unknown variable is $\mathbf{t}$. It means that here the compliance model does not allow us to compute the modified target point $\mathbf{t}_{\text{0}}^{\text{(F)}}$ straightforwardly, while the linear compensation technique directly operates with Cartesian compliance matrix [5]. 

Since ${{\mathbf{t}}_{0}}$ and $\mathbf{t}_{\text{0}}^{\text{(F)}}$ are close enough, to  solve equation (2) for $\mathbf{t}_{\text{0}}^{\text{(F)}}$, the Newton-Raphson technique can be applied. It yields the following iterative scheme

\begin{equation}\label{Eq:3}
\mathbf{t}{{_{\text{0}}^{\text{(F)}}}^{\prime}}=\mathbf{t}_{\text{0}}^{\text{(F)}}+\mathbf{K}_{\text{t}\text{.p}\text{.}}^{-1}({{\mathbf{t}}_{0}}|\mathbf{t}_{\text{0}}^{\text{(F)}})\left( \mathbf{F}-f({{\mathbf{t}}_{0}}|\mathbf{t}_{\text{0}}^{\text{(F)}}) \right)
\end{equation}	
where the prime corresponds to the next iteration and $\mathbf{K}_{\text{t}\text{.p}\text{.}}^{{}}({{\mathbf{t}}_{0}}|\mathbf{t}_{\text{0}}^{\text{(F)}})$ is the stiffness matrix computed with respect to the second argument of the function $\mathbf{F}=f\left( \mathbf{t}|\mathbf{t}_{\text{0}}^{{}} \right)$ at the original target point (i.e. for $\mathbf{t}={{\mathbf{t}}_{0}}$) assuming that unloaded configuration is modified and corresponds to the end-effector location $\mathbf{t}_{\text{0}}^{\text{(F)}}$. Here $\mathbf{F}$ stands for the solution of equation (2), while the function $f\left( {{\mathbf{t}}_{0}}|\mathbf{t}_{\text{0}}^{\text{(F)}} \right)$ defines the loading for the current end-effector location under the loading $\mathbf{t}_{\text{0}}^{\text{(F)}}$. 

To overcome computational difficulties related to the evaluation of the matrix $\mathbf{K}_{\text{t}\text{.p}\text{.}}^{{}}({{\mathbf{t}}_{0}}|\mathbf{t}_{\text{0}}^{\text{(F)}})$, it is possible to use its simple approximation that does not change from iteration to iteration. In particular, assuming that $\mathbf{t}$ and ${{\mathbf{t}}_{0}}$ are close enough and the stiffness properties do not vary substantially in their neighborhood, the stiffness model (2) can be approximated by a linear expression $\mathbf{F}={{\mathbf{K}}_{\text{C}}}(\mathbf{t}-{{\mathbf{t}}_{0}})$, which includes the conventional Cartesian stiffness matrix $\mathbf{K}_{\text{C}}^{{}}$. This allows us to replace the above derivative matrix $\mathbf{K}_{\text{t}\text{.p}\text{.}}^{{}}$ by  $-\mathbf{K}_{\text{C}}^{{}}$ and to present the iterative scheme (3) as 

\begin{equation}\label{Eq:4}
\mathbf{t}{{_{\text{0}}^{\text{(F)}}}^{\prime }}=\mathbf{t}_{\text{0}}^{\text{(F)}}-\alpha \mathbf{K}_{\text{C}}^{-1}({{\mathbf{t}}_{0}}|\mathbf{t}_{\text{0}}^{\text{(F)}})\left(\mathbf{F}-f({{\mathbf{t}}_{0}}|\mathbf{t} _{\text{0}}^{\text{(F)}}) \right)
\end{equation}
where $\alpha \in (0,1)$ is the scalar parameter ensuring the convergence. Using the non-linear compliance model (1), this idea can also be implemented in an iterative algorithm 

\begin{equation}\label{Eq:5}
	\mathbf{t}{{_{\text{0}}^{\text{(F)}}}^{\prime }}=\mathbf{t}_{\text{0}}^{\text{(F)}}+\alpha \left({{\mathbf{t}}_{0}}-{{f}^{-1}}(\mathbf{F}|\mathbf{t}_{\text{0}}^{\text{(F)}}) \right)
\end{equation}
which does not include stiffness matrices $\mathbf{K}_{\text{C}}^{{}}$ or $\mathbf{K}_{\text{t}\text{.p}\text{.}}^{{}}$. Obviously, this is the most computationally convenient solution and it will be used in the next section. 

\begin{figure}[b]
\center
\includegraphics{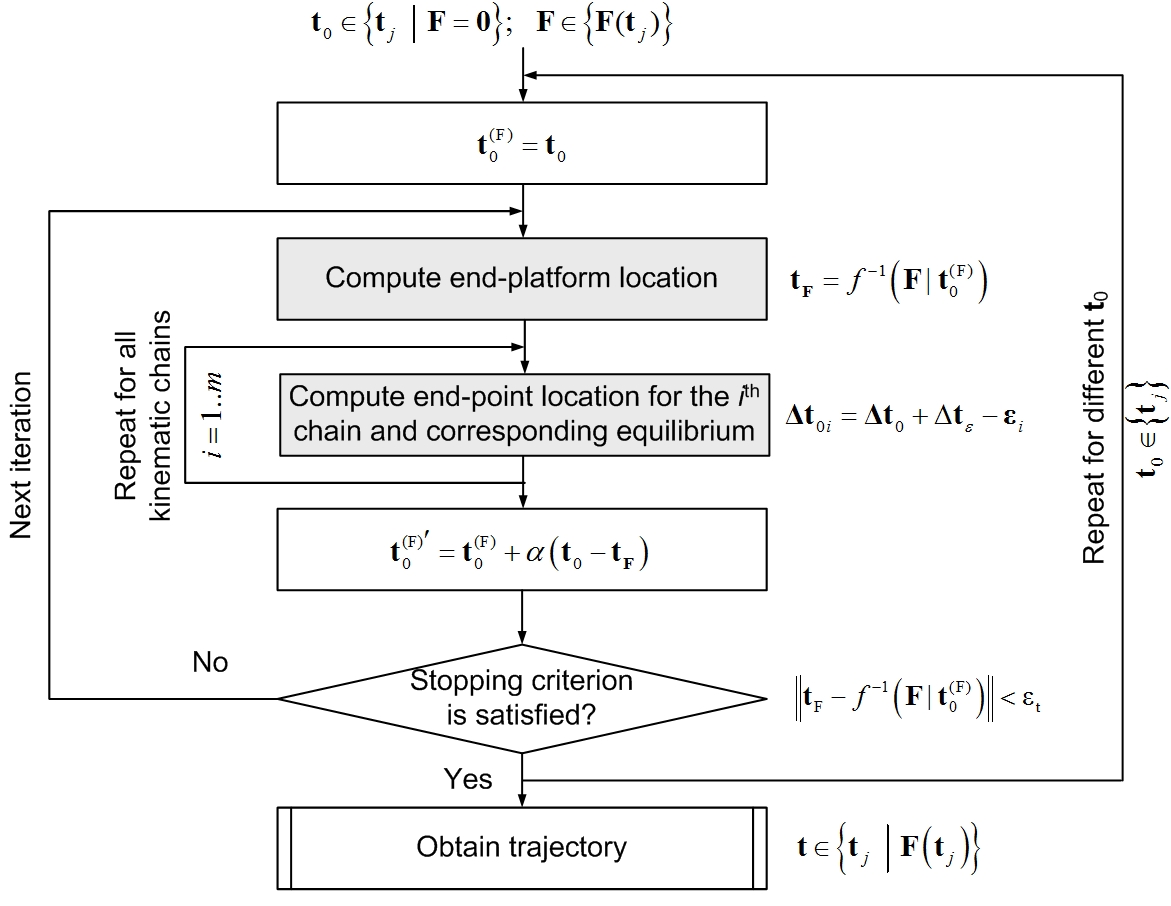}
\caption{Procedure for compensation of compliance errors in parallel manipulator.}
\label{Figure:2}
\end{figure}

It should be mentioned that the considered case deals with a perfect parallel manipulator where end-points of all kinematic chains are aligned and matched. However, in practice, kinematic chains may include some errors that do not allow us to assemble them in a parallel manipulator with the same end-effector location. In this case it is required to compensate two types of errors (caused by the external loading $\mathbf{F}$ and inaccuracy in the serial chains). The second source of errors can be taken into account by changing of target location $\Delta{{\mathbf{t}}_{0i}}$ for each kinematic chain

\begin{equation}\label{Eq:6}
\Delta{{\mathbf{t}}_{0i}}=\Delta{{\mathbf{t}}_{0}}+\Delta {{\mathbf{t}}_{\varepsilon }}-{{\mathbf{\varepsilon }}_{i}}
\end{equation}
where $\Delta {{\mathbf{t}}_{\varepsilon }}$ is the end-effector deflections due to assembling of non-perfect kinematic chains and ${{\mathbf{\varepsilon }}_{i}}$ is shifting of the end-point location of ith kinematic chain because of geometrical errors. Using the principle of virtual work it can be proved that $\Delta {{\mathbf{t}}_{\varepsilon }}$ can be computed as 

\begin{equation}\label{Eq:7}
\Delta {{\mathbf{t}}_{\varepsilon }}={{\left( \sum\limits_{i=1}^{m}{\mathbf{K}_{\text{C}}^{(i)}} \right)}^{-1}}\sum\limits_{i=1}^{m}{\left( \mathbf{K}_{\text{C}}^{(i)}{{\mathbf{\varepsilon }}_{i}} \right)}
\end{equation}
where $\mathbf{K}_{\text{C}}^{(i)}$ defines the Cartesian stiffness matrix of i-th kinematic chain that can be computed using techniques proposed in [2] and $m$ is the number of kinematic chains in the parallel manipulator. More detailed presentation of the developed iterative routines is given in Fig. 2.

Hence, using the proposed computational techniques, it is possible to compensate the essential compliance errors by proper adjusting the reference trajectory that is used as an input for robotic controller. In this case, the control is based on the inverse kinetostatic model (instead of kinematic one) that takes into account both the manipulator geometry and elastic properties of its links and joints. Efficiency of this technique is confirmed by an example presented in the next section.

\section{Illustrative example: compliance error compensation for milling} 
Let us illustrate the compliance errors compensation technique by an example of the circle groove milling with Orthoglide manipulator (Fig. 3). Detailed specification of this manipulator can be founded in [6]. According to [7], such technological process causes the loading ${{F}_{r}}=215\,N$; ${{F}_{t}}=-10\,N$; ${{F}_{z}}=-25\,N$ that together with angular parameter $\varphi =[0,{{360}^{\circ }}]$ define the forces ${{F}_{x}}$ and ${{F}_{y}}$ (Fig. 3b,c). Here, the tool length $h$ is equal to $100\,mm$. It is assumed that the manipulator has two sources of inaccuracy:

\begin{figure}[b]
\center
\includegraphics{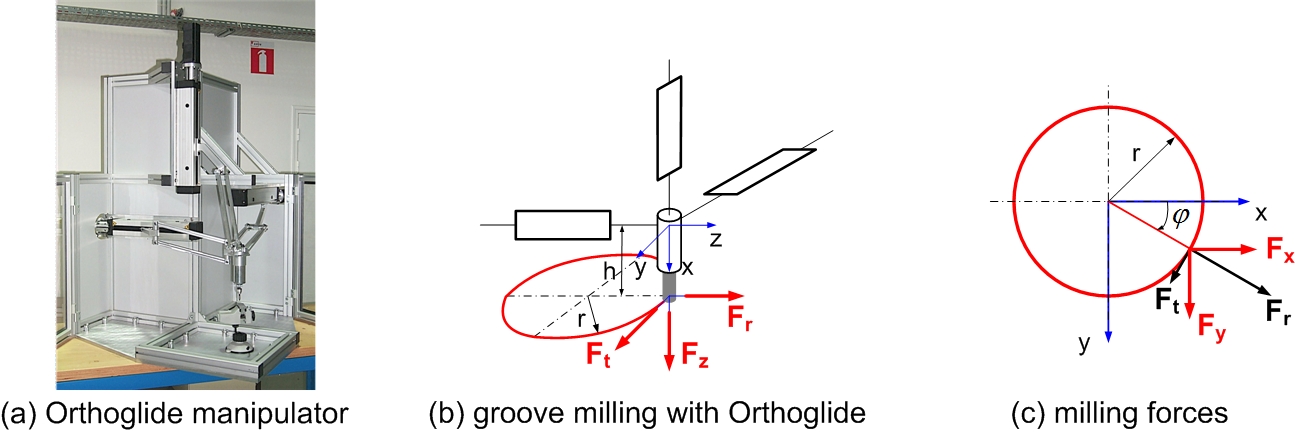}
\caption{Milling forces and trajectory location for groove milling using Orthoglide manipulator.}
\label{Figure:3}
\end{figure}

\begin{enumerate}
	\item the assembling errors in the kinematic chains (assembling errors in actuator angular locations of about $1^{\circ}$ around the corresponding actuated axis) causing internal forces and relevant deflections in joints and links;
 \item the external loading $\left\| \mathbf{F} \right\|=217\,N$ which generates essential compliance deflections causing non-desirable end-platform displacement.
\end{enumerate}

In order to illustrate influence of different error sources on the machining trajectory, let us focus on the 1 mm radius of the circle that should be machined. In this case, the stiffness matrix is almost the same along the trajectory. Modeling results for the neighborhood of point $Q_1$ (see [2] for details) are presented in Fig. 4. They show the influence of different error sources on the machining trajectory without compensation and the revised machining trajectory that should be implemented in robot controller in order to follow the target trajectory while machining. Here, path 5 compensates the effects seen in path 4 such that circle 1 is achieved. It can be seen that the center of path 5 is on the opposite side of circle 1 compared to path 4. It can also be seen that the main elliptic direction in path 4 becomes the smallest elliptic direction in path 5. It should be mentioned that because of the torque induced by the cutting forces (tool length 100 mm), the target trajectory and shifted trajectory under the cutting forces are intersecting.
 
\begin{figure}[hbtp]
    \begin{tabular}{cc}
       \begin{minipage}[c]{0.5 \linewidth}
           \includegraphics{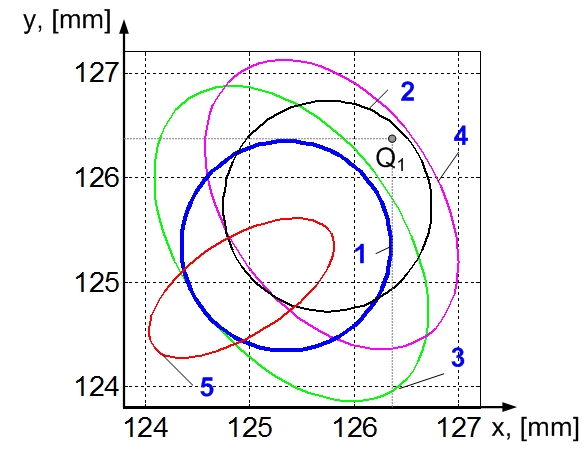}      
       \end{minipage} 
       \begin{minipage}[c]{0.5 \linewidth}
          \begin{enumerate}
           \item[(1)] Target trajectory; 
           \vspace{1.5mm}
           \item[(2)] Shifting of target trajectory caused by errors in serial chains  (assembling errors); 
           \vspace{1.5mm}
           \item[(3)] Shifting of target trajectory caused by cutting force (compliance errors); 
           \vspace{1.5mm}
           \item[(4)] Shifting of target trajectory caused by cutting force and errors in serial chains; 
           \vspace{1.5mm}
           \item[(5)] Adjusted trajectory, that insure following the target trajectory while machining.
          \end{enumerate} 
      \end{minipage}
    \end{tabular}
    \caption{Influence of different error sources on the machining trajectory.}    
    \protect\label{Figure:4}
\end{figure}

\begin{figure}[t]
\center
\includegraphics{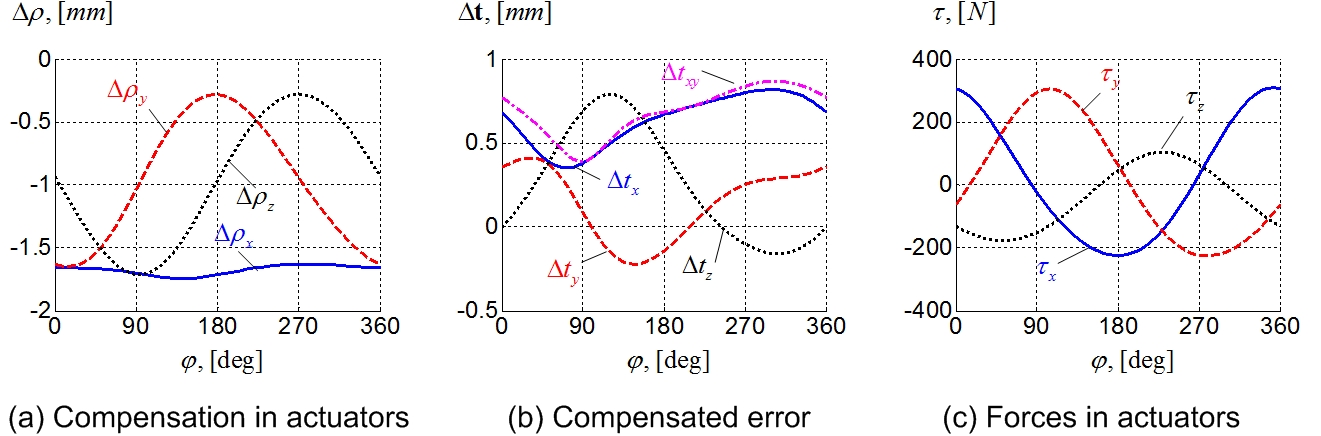}
\caption{Compliance error compensation for Orthoglide milling application.}
\label{Figure:5}
\end{figure}

Figure 5 presents results for the milling of the 50 mm circle. In this case, without compensation, the compliance errors can exceed 0.8 mm. After compensation, the above mentioned errors are reduced to zero (it is obvious that in practice, the compensation level is limited by the accuracy of the stiffness model). This compensation is achieved due to the modification of the actuator coordinates $\mathbf{\rho }$ along the machining trajectory. Compared to the relevant values computed via the inverse kinematics, the actuator coordinates differ up to 1.7 mm. Corresponding forces in actuators can reach 300 N. Some more results on the compliance errors compensation are presented in Fig. 5, which includes plots showing modifications of the actuator coordinates $\Delta \mathbf{\rho }$, values of compensated end-effector displacement $\Delta \mathbf{t}$ and the torques in actuators $\tau$. It should be mentioned that while implementing target trajectory in the robot controller additional control errors may arise.

Hence, the developed algorithm demonstrates good convergence. It is able to compensate the compliance errors and can be efficient both for off-line trajectory planning and for on-line errors compensation.

\section{Conclusions}
The paper presents a new technique for on-line and off-line compensation of the compliance errors caused by external loadings in parallel manipulators (including over-constrained ones) composed of both perfect and non-perfect serial chains. In contrast to previous works this technique is based on nonlinear stiffness model (inverse kinetostatic model) that gives essential benefits for robotic-based machining, where the elastic deflections can be essential. The advantages and practical significance are illustrated by groove milling with Orthoglide manipulator. 

\begin{acknowledgement}
The work presented in this paper was partially funded by the Region ``Pays de la Loire", France and by the project  ANR COROUSSO, France.\end{acknowledgement}

\end{document}